\documentclass[10pt,twocolumn,letterpaper]{article}

\usepackage{iccv}
\usepackage{times}
\usepackage{bbm}
\usepackage{epsfig}
\usepackage{graphicx}
\usepackage{amsmath}
\usepackage{amssymb}
\usepackage{color}
\usepackage[ruled]{algorithm2e}
\usepackage{booktabs}
\usepackage{multirow}

\usepackage[pagebackref=true,breaklinks=true,letterpaper=true,colorlinks,bookmarks=false]{hyperref}

\iccvfinalcopy 

\def\iccvPaperID{1865} 

\ificcvfinal\fi
\begin{document}


\title{Multinomial Distribution Learning for Effective Neural Architecture Search}

\author{
    Xiawu Zheng $^{1,2}$\ ,
    Rongrong Ji$^{1,2}$\thanks{Corresponding Author.} ,
    Lang Tang$^{1,2}$,
    Baochang Zhang$^3$,
    Jianzhuang Liu$^4$,
    Qi Tian$^5$ \\
    $^1$Media Analytics and Computing Lab, Department of Artificial Intelligence, \\
    School of Informatics, Xiamen University, 361005, China,\\
    $^2$Peng Cheng Laboratory, Shenzhen, China,
    $^3$Beihang University, China,\\
    $^4$Huawei Noah's Ark Lab, $^5$Department of Computer Science, University of Texas at San Antonio    \\
    {\tt\small \{zhengxiawu,langt\}@stu.xmu.edu.cn, rrji@xmu.edu.cn}\\
    {\tt\small bczhang@buaa.edu.cn, liu.jianzhuang@huawei.com, qitian@cs.utsa.edu}
}
\maketitle

\begin{abstract}
Architectures obtained by Neural Architecture Search (NAS) have achieved highly competitive performance in various computer vision tasks. However, the prohibitive computation demand of forward-backward propagation in deep neural networks and searching algorithms makes it difficult to apply NAS in practice. In this paper, we propose a Multinomial Distribution Learning for extremely effective NAS, which considers the search space as a joint multinomial distribution, i.e., the operation between two nodes is sampled from this distribution, and the optimal network structure is obtained by the operations with the most likely probability in this distribution. Therefore, NAS can be transformed to a multinomial distribution learning problem, i.e., the distribution is optimized to have high expectation of the performance. Besides, a hypothesis that the performance ranking is consistent in every training epoch is proposed and demonstrated to further accelerate the learning process. Experiments on CIFAR-10 and ImageNet demonstrate the effectiveness of our method. On CIFAR-10, the structure searched by our method achieves 2.55\% test error, while being 6.0$\times$ (only 4 GPU hours on GTX1080Ti) faster compared with state-of-the-art NAS algorithms. On ImageNet, our model achieves 75.2\% top-1 accuracy under MobileNet settings (MobileNet V1/V2), while being 1.2$\times$ faster with measured GPU latency. Test code with pre-trained models are available at \url{ https://github.com/tanglang96/MDENAS}
\end{abstract}

\section{Introduction}\label{sec:intro}
Given a dataset, Neural architecture search (NAS) aims to discover high-performance convolution architectures with a searching algorithm in a tremendous search space. NAS has achieved much success in automated architecture engineering for various deep learning tasks, such as image classification \cite{liu2018progressive, zoph2018learning, zheng2019dynamic}, language modeling \cite{liu2018darts, zoph2016neural} and semantic segmentation \cite{liu2019auto,chen2018searching}. As mentioned in \cite{elsken2018neural}, NAS methods consist of three parts: search space, search strategy, and performance estimation. A conventional NAS algorithm samples a specific convolutional architecture by a search strategy and estimates the performance, which can be regarded as an objective to update the search strategy. Despite the remarkable progress, conventional NAS methods are prohibited by intensive computation and memory costs. For example, the reinforcement learning (RL) method in \cite{zoph2018learning} trains and evaluates more than 20,000 neural networks across 500 GPUs over 4 days. Recent work in \cite{liu2018darts} improves the scalability by formulating the task in a differentiable manner where the search space is relaxed to a continuous space, so that the architecture can be optimized with the performance on a validation set by gradient descent. However, differentiable NAS still suffers from the issued of high GPU memory consumption, which grows linearly with the size of the candidate search set.
\begin{figure*}[t]
\begin{center}
\includegraphics[width=1.0\linewidth]{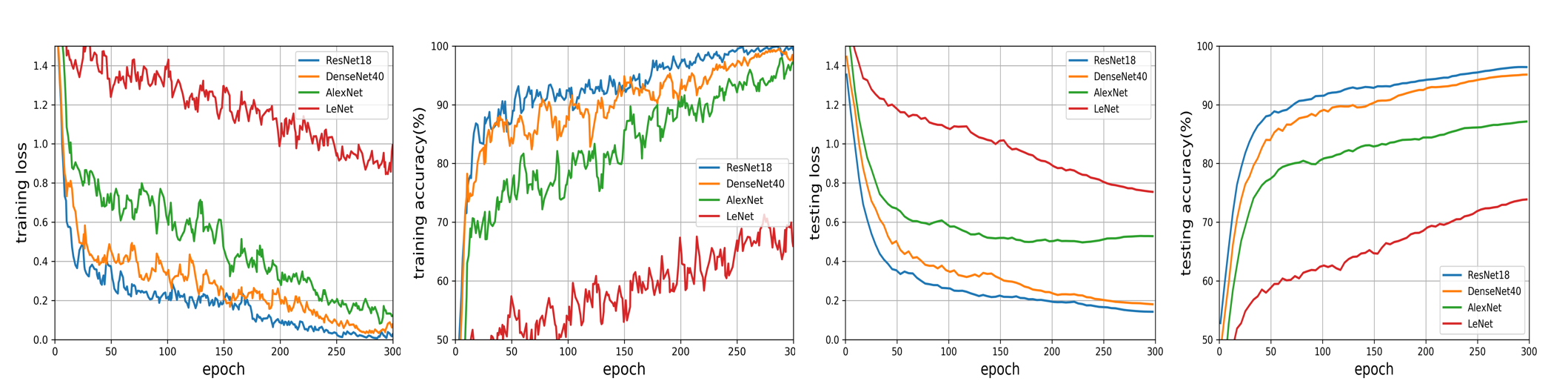}
\end{center}
   \caption{We randomly choose widely used LeNet \cite{lecun1998gradient}, AlexNet \cite{krizhevsky2010convolutional}, ResNet-18\cite{he2016deep} and DenseNet-BC($k=40$) \cite{huang2017densely} to illustrate the proposed \emph{Performance Ranking Hypothesis}. The training and testing are conducted on CIFAR-10. We report the top1 error and loss learning curves on both training and testing set. As we can see in the figure, the \emph{ranking} of the test loss and accuracy keeps consistent in every training epoch, \emph{i.e.,} a good architecture tends to have better performance in the whole training process. }
\label{fig:performance_ranking_hyp}
\end{figure*}

 Indeed, most NAS methods \cite{zoph2018learning, liu2019auto} perform the performance estimation using standard training and validation over each searched architecture, typically, the architecture has to be trained to converge to get the final evaluation on validation set, which is computationally expensive and limits the search exploration. However, if the evaluation of different architectures can be ranked within a few epochs, why do we need to estimate the performance after the neural network converges? Consider an example in Fig.~\ref{fig:performance_ranking_hyp}, we randomly sample different architectures (LeNet \cite{lecun1998gradient}, AlexNet \cite{krizhevsky2012imagenet}, ResNet-18 \cite{he2016deep} and DenseNet \cite{huang2017densely}) with different layers, the performance ranking in the training and testing is consistent (\emph{i.e}, the performance ranking is ResNet-18 $>$ DenseNet-BC $>$ AlexNet $>$ LeNet on different networks and training epochs). Based on this observation, we state the following hypothesis for performance ranking: 

\noindent \textbf{Performance Ranking Hypothesis.} \emph{If Cell A has higher validation performance than Cell B on a specific network and a training epoch, Cell A tends to be better than Cell B on different networks after the trainings of these netwoks converge.}

\noindent Here, a cell is a fully convolutional directed acyclic graph (DAG) that maps an input tensor to an output tensor, and the final network is obtained through stacking different numbers of cells, the details of which are described in Sec.~\ref{sec:3}. 

The hypothesis illustrates a simple yet important rule in neural architecture search. The comparison of different architectures can be finished at early stages, as the ranking of different architectures is sufficient, whereas the final results are unnecessary and time-consuming. Based on this hypothesis, we propose a simple yet effective solution to neural architecture search, termed as Multinomial distribution for efficient Neural Architecture Search (MdeNAS), which directly formulates NAS as a distribution learning process. Specifically, the probabilities of operation candidates between two nodes are initialized equally, which can be considered as a multinomial distribution. In the learning procedure, the parameters of the distribution are updated through the current performance in every epoch, such that the probability of a bad operation is transferred to better operations. With this search strategy, MdeNAS is able to fast and effectively discover high-performance architectures with complex graph topologies within a rich search space.

In our experiments, the convolutional cells designed by MdeNAS achieve strong quantitative results. The searched model reaches 2.55\% test error on CIFAR-10 with less parameters. On ImageNet, our model achieves 75.2\% top-1 accuracy under MobileNet settings (MobileNet V1/V2  \cite{howard2017mobilenets, sandler2018mobilenetv2}), while being 1.2$\times$ faster with measured GPU latency. The contributions of this paper are summarized as follows:
\begin{itemize}
\item We introduce a novel algorithm for network architecture search, which is applicable to various large-scale datasets as the memory and computation costs are similar to common neural network training.
\item We propose a performance ranking hypothesis, which can be incorporated into the existing NAS algorithms to speed up its search.
\item The proposed method achieves remarkable search efficiency, \emph{e.g.}, 2.55\% test error on CIFAR-10 in 4 hours with 1 GTX1080Ti (6.0$\times$ faster compared with state-of-the-art algorithms), which is attributed to using our distribution learning that is entirely different from RL-based \cite{baker2016designing,zoph2018learning} methods and differentiable methods \cite{liu2018darts, xie2018snas}.
\end{itemize}

\section{Related Work}
As first proposed in \cite{zoph2016neural, zoph2018learning}, automatic neural network search in a predefined architecture space has received significant attention in the last few years. To this end, many search algorithms have been proposed to find optimal
architectures using specific search strategies. Since most hand-crafted CNNs are built by stacked \emph{reduction} (\emph{i.e.}, the spatial dimension of the input is reduced) and \emph{norm} (\emph{i.e.} the spatial dimensionality of the input is preserved) cells \cite{huang2017densely, he2016deep, hu2018squeeze}, the works in \cite{zoph2016neural, zoph2018learning} proposed to search networks under the same setting to reduce the search space.
The works in \cite{zoph2016neural, zoph2018learning,baker2016designing} use reinforcement learning as a meta-controller, to explore the architecture search space. The works in \cite{zoph2016neural, zoph2018learning} employ a recurrent neural network (RNN) as the policy to sequentially sample a string encoding a specific neural architecture. The policy network can be trained with the policy gradient algorithm or the proximal policy optimization. The works in \cite{cai2018efficient, cai2018path, liu2018progressive} regard the architecture search space as a tree structure for network transformation, \emph{i.e.}, the network is generated by a farther network with some predefined operations, which reduces the search space and speeds up the search. An alternative to RL-based methods is the \emph{evolutionary} approach, which optimizes the neural architecture by evolutionary algorithms \cite{xie2017genetic, real2018regularized}. 

However, the above architecture search algorithms are still computation-intensive. Therefore some recent works are proposed to accelerate NAS by \emph{one-shot} setting, where the network is sampled by a hyper representation graph, and the search process can be accelerated by parameter sharing \cite{pham2018efficient}. For instance, DARTS \cite{liu2018darts} optimizes the weights within two node in the hyper-graph jointly with a continuous relaxation. Therefore, the parameters can be updated via standard gradient descend. However, \emph{one-shot} methods suffer from the issue of large GPU memory consumption. To solve this problem, ProxylessNAS \cite{cai2018proxylessnas} explores the search space without a specific agent with path binarization \cite{courbariaux2015binaryconnect}. However, since the search procedure of ProxylessNAS is still within the framework of \emph{one-shot} methods, it may have the same complexity, \emph{i.e.,} the benefit gained in ProxylessNAS is a \emph{trade-off} between exploration and exploitation. That is to say, more epochs are needed in the search procedure. Moreover, the search algorithm in \cite{cai2018proxylessnas} is similar to previous work, either differential or RL based methods \cite{liu2018darts, zoph2018learning}.

Different from the previous methods, we encode the path/operation selection as a distribution sampling, and achieve the optimization of the controller/proxy via distribution learning. Our learning process further integrates the proposed hypothesis to estimate the merit of each operation/path, which achieves an extremely efficient NAS search.
\begin{figure}
\begin{center}
\includegraphics[width=1.0\linewidth]{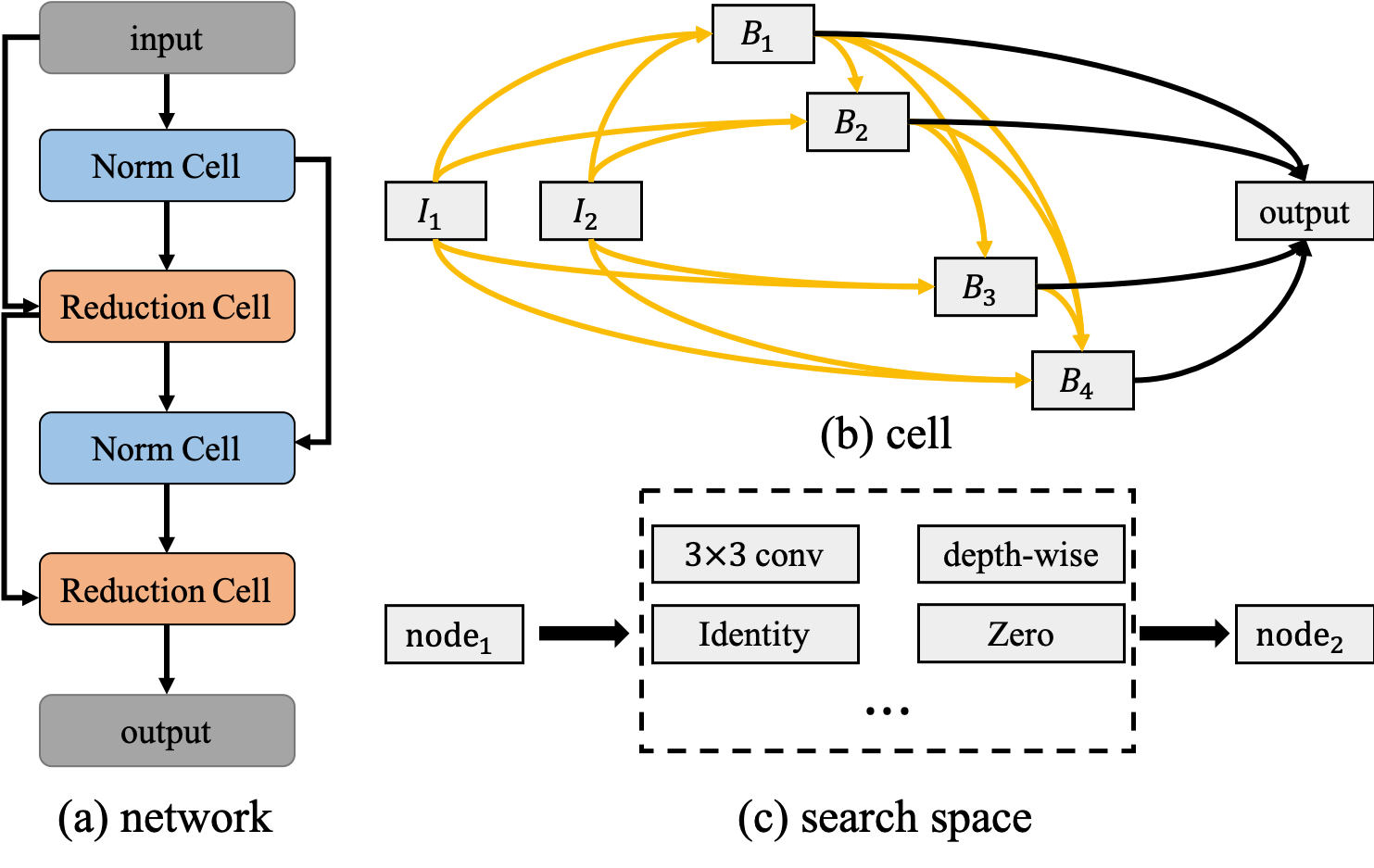}
\end{center}
   \caption{Searching networks with different scales. (a) A network consists of stacked cells, and each cell takes the output of two previous cells as input. (b) A cell contains 7 nodes, two input nodes $I_1$ and $I_2$, four intermediate nodes $B_1, B_2, B_3, B_4$ that apply sampled operations on the input nodes and upper nodes, and an output node that concatenates the outputs of the four intermediate nodes. (c) The edge between two nodes denotes a possible operation according to a multinomial distribution in the search space.}
\label{fig:search_space}
\end{figure}
\begin{figure*}
\begin{center}
\includegraphics[width=1.0\linewidth]{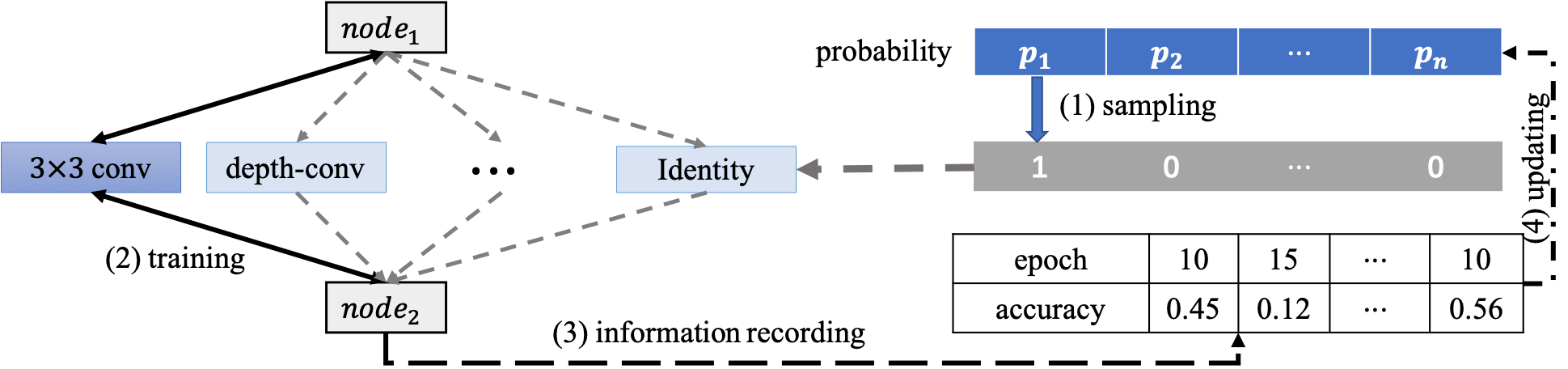}
\end{center}
   \caption{The overall search algorithm: (1) Sample one operation in the search space according to the corresponding multinomial distribution with parameters $\theta$. (2) Train the generated network with one forward and backward propagation. (3) Test the network on the validation set and record the feedback (epoch and accuracy). (4) Update the distribution parameters according to the proposed distribution learning algorithm. In the right table, the epoch number of operation 1 is 10, which means that this operation is selected 10 times among all the epochs.}
\label{fig:search_alg}
\end{figure*}
\section{Architecture Search Space}\label{sec:3}
In this section, we describe the  architecture search space and the method to build the network. We follow the same settings as in previous NAS works \cite{liu2018darts, liu2018progressive,zoph2018learning} to keep the consistency. As illustrated in Fig.~\ref{fig:search_space}, the network is defined in different \emph{scales}: network, cell, and node.
\subsection{Node}\label{sec:3.1}
Nodes are the fundamental elements that compose cells. Each node $x^i$ is a specific tensor (\emph{e.g.}, a feature map in convolutional neural networks) and each directed edge $(i,j)$ denotes an operation $o^{(i,j)}$ sampled from the operation search space to transform node $x^i$ to another node $x^j$, as illustrated in Fig.~\ref{fig:search_space}{\color{red}(c)}. There are three types of nodes in a cell: input node $x_I$, intermediate node $x_B$, and output node $x_O$. Each cell takes the previous output tensor as an input node, and generates the intermediate nodes $x_B^i$ by applying sampled operations $o^{(i,j)}$ to the previous nodes  ($x_I$ and $x_B^j, j \in [1,i)$). The concatenation of all intermediate nodes is regarded as the final output node.

Following \cite{liu2018darts} set of possible operations, denoted as $\mathcal{O}$, consists of the following 8 operations: (1) $3\times3$ max pooling. (2) no connection (zero). (3) $3\times3$ average pooling. (4) skip connection (identity). (5) $3\times3$ dilated convolution with rate 2. (6) $5\times5$ dilated convolution with rate 2. (7) $3\times3$ depth-wise separable convolution. (8) $5\times5$ depth-wise separable convolution.

We simply employ element-wise addition at the input of a node with multiple operations (edges). For example, in Fig.~\ref{fig:search_space}{\color{red}(b)}, $B_2$ has three operations, the results of which are added element-wise and then considered as $B_2$.

\subsection{Cell}
A cell is defined as a tiny convolutional network mapping an $H\times W\times F$ tensor to another $H^{\prime} \times W^{\prime} \times F^{\prime}$. There are two types of cells, norm cell and reduction cell. A norm cell uses the operations with stride 1, and therefore $H^{\prime} = H$ and $W^{\prime} = W$. A reduction cell uses the operations with stride 2, so  $H^{\prime} = H/2$ and $W^{\prime} = W/2$. For the numbers of filters $F$ and $F^{\prime}$, a common heuristic in most human designed convolutional neural networks \cite{he2016deep,huang2017densely,krizhevsky2012imagenet,simonyan2014very, fan2019shifting, zhao2019contrast} is to double $F$ whenever the spatial feature map is halved. Therefore, $F^{\prime} = F$ for stride 1, and $F^{\prime}=2F$ for stride 2. 

As illustrated in Fig.~\ref{fig:search_space}{\color{red}(b)}, the cell is represented by a DAG with 7 nodes (two input nodes $I_1$ and $I_2$, four intermediate nodes $B_1, B_2, B_3, B_4$ that apply sampled operations on the input and upper nodes, and an output node that concatenates the intermediate nodes). The edge between two nodes denote a possible operation according to a multinomial distribution $p_{(node_1, node_2)}$ in the search space. In training, the input of an intermediate node is obtained by element-wise addition when it has multiple edges (operations). In testing, we select the top \emph{K} probabilities to generate the final cells. Therefore, the size of the whole search space is $2 \times 8^{|\mathcal{E_N}|}$, where $\mathcal{E_N}$ is the set of possible edges with $N$ intermediate nodes. In our case with $N=4$, the total number of cell structures is $2 \times 8^{2+3+4+5} = 2 \times 8^{14}$, which is an extremely large space to search, and thus requires efficient optimization methods.

\subsection{Network}
As illustrated in Fig.~\ref{fig:search_space}{\color{red}(a)}, a network consists of a predefined number of stacked cells, which can be either norm cells or reduction cells each taking the output of two previous cells as input. At the top of the network, global average pooling followed by a softmax layer is used for final output. Based on the \emph{Performance Ranking Hypothesis}, we train a small (\emph{e.g.}, 6 layers) stacked model on the relevant dataset to search for norm and reduction cells, and then generate a deeper network (\emph{e.g.}, 20 layers) for evaluation.
The overall CNN construction process and the search space are identical to \cite{liu2018darts}. But note that our search algorithm is different.

\section{Methodology}
In this section, our NAS method is presented. We first describe how to sample the network mentioned in Sec.~\ref{sec:3} to reduce GPU memory consumption during training. Then, we present a multinomial distribution learning to effectively optimize the distribution parameters using the proposed hypothesis.

\subsection{Sampling}
As mentioned in Sec.~\ref{sec:3.1}, the diversity of network structures is generated by different selections of \emph{M} possible paths (in this work, $M=8$) for every two nodes.  Here we initialize the probabilities of these paths as $p_i = \frac{1}{M}$ in the beginning for exploration. In the sampling stage, we follow the work in \cite{cai2018proxylessnas} and transform the \emph{M} real-valued probabilities $\{p_i\}$ with binary gates $\{g_i\}$:
\begin{equation}\label{eq:prob}
g = 
\left\{
             \begin{array}{lr}
             \underbrace{[1,0, ...,0]}_{M}  \, {\rm with} \, {\rm probability} \, p_1\\
             \quad \quad ...\\
             \underbrace{[0,0, ...,1]}_{M} \, {\rm with} \, {\rm probability} \, p_M
             \end{array}
\right.
\end{equation}
The final operation between nodes $i$ and $j$ is obtained by:
\begin{equation}\label{eq:gate}
o^{(i,j)} = o^{(i,j)} * g = 
\left\{
             \begin{array}{lr}
             o_1 \, {\rm with} \, {\rm probability} \, p_1\\
             ...\\
             o_M \, {\rm with} \, {\rm probability} \, p_M.
             \end{array}
\right.
\end{equation}
As illustrated in the previous equations, we sample only one operation at run-time, which effectively reduces the memory cost compared with \cite{liu2018darts}.

\subsection{Multinomial Distribution Learning}
Previous NAS methods are time and memory consuming. The use of reinforcement learning further prohibits the methods with the delay reward in network training, \emph{i.e.,} the evaluation of a structure is usually finished after the network training converges. On the other hand, as mentioned in Sec.~\ref{sec:intro}, according to the \emph{Performance Ranking Hypothesis}, we can perform the evaluation of a cell when training the network. As illustrated in Fig.~\ref{fig:search_alg}, the training epochs and accuracy for every operation in the search space are recorded. Operations $A$ is better than $B$, if operation $A$ has fewer training epochs and higher accuracy.

Formally, for a specific edge between two nodes, we define the operation probability as $p$, the training epoch as $\mathcal{H}^e$, and the accuracy as $\mathcal{H}^a$, each of which is a real-valued column vector of length $M=8$. To clearly illustrate our learning method, we further define the differential of epoch as:
\begin{equation}\label{eq:delta_epoch}
\centering
\Delta  \mathcal{H}^e= 
\left[
             \begin{array}{lr}
             (\vec{1} \times  \mathcal{H}^e_1 -  \mathcal{H}^e )^{T}\\
              \quad \quad \quad ...\\              
              ( \vec{1} \times  \mathcal{H}^e_M -  \mathcal{H}^e )^{T}
             \end{array}
\right],
\end{equation}
and the differential of accuracy as:
\begin{equation}\label{eq:delta_ac}
\Delta  \mathcal{H}^a= 
\left[
             \begin{array}{lr}
             (\vec{1} \times  \mathcal{H}^a_1 -  \mathcal{H}^a)^T\\
             \quad \quad \quad ...\\
             ( \vec{1} \times  \mathcal{H}^a_M -  \mathcal{H}^a)^T
             \end{array}
\right],
\end{equation}
where $\vec{1}$ is a column vector with length 8 and all its elements being $1$,  $\Delta  \mathcal{H}^e$ and $\Delta  \mathcal{H}^a$ are $8 \times 8$ matrices, where $\Delta  \mathcal{H}^e_{i,j} = \mathcal{H}^e_i-\mathcal{H}^e_j, \Delta  \mathcal{H}^a_{i,j} = \mathcal{H}^a_i-\mathcal{H}^a_j $. After one epoch training, the corresponding variables $\mathcal{H}^e$, $\mathcal{H}^a$, $\Delta  \mathcal{H}^e$ and $\Delta  \mathcal{H}^a$ are calculated by the evaluation results. The parameters of the multinomial distribution can be updated through:
\begin{equation}\label{eq:alg}
\begin{aligned}
p_i \leftarrow p_i + \alpha *  (&\sum_j \mathbbm{1}(\Delta  \mathcal{H}^e_{i,j} < 0 , \Delta  \mathcal{H}^a_{i,j} > 0) - 
\\
&\sum_j \mathbbm{1}(\Delta  \mathcal{H}^e_{i,j} > 0 , \Delta  \mathcal{H}^a_{i,j} < 0)),
\end{aligned}
\end{equation}
where $\alpha$ is a hyper-parameter, and $\mathbbm{1}$ denotes as the indicator function that equals to one if its condition is true.

As we can see in Eq.~\ref{eq:alg}, the probability of a specific operation $i$ is enhanced with fewer epochs ($\Delta  \mathcal{H}^e_{i,j} < 0$) and higher performance ($\Delta  \mathcal{H}^a_{i,j} > 0$). At the same time, the probability is reduced with more epochs ($\Delta  \mathcal{H}^e_{i,j} > 0$) and lower performance ($\Delta  \mathcal{H}^a_{i,j} < 0$). Since Eq.~\ref{eq:alg} is applied after every training epoch, the probability in the search space can be effectively converge and stabilize after a few epochs. Together with the proposed performance ranking hypothesis (demonstrated latter in Section~\ref{experiment}), our multinomial distribution learning algorithm for NAS is extremely efficient, and achieves a better performance compared with other state-of-the-art methods under the same settings. Considering the performance ranking is consisted of different layers according to the hypothesis, to further improve the search efficiency, we replace the search network in \cite{liu2018darts} with another shallower one (only 6 layers), which takes only 4 GPU hours of searching on CIFAR-10.

To generate the final network, we first select the operations with highest probabilities in all edges. For nodes with multi-input, we employ element-wise addition with top $K$ probabilities. The final network consists of a predefined number of stacked cells, using either norm or reduction cells. Our multinomial distribution learning algorithm is presented in Alg.~\ref{alg:MDL}.

\begin{algorithm}[t]
\caption{Multinomial Distribution Learning \label{alg:MDL}}
\LinesNumbered
\KwIn{Training data: $\mathcal{D}_t$; Validation data: $\mathcal{D}_v$; CNN model: $\mathcal{F}$}.
\KwOut{Cell operation probabilities: $\mathcal{P}$}.\
\For{t= \rm 1,...,T \bfseries epoch}{
Sample the operation according to Equation~\ref{eq:prob}\;
Train the network with 1 epoch\;
Validate the network on $\mathcal{D}_v$\;
Caculate the differential of epoch and accuracy according to Equation~\ref{eq:delta_epoch} and Equation~\ref{eq:delta_ac}\;
Update the probabilities with Equation~\ref{eq:alg}\;
}
\end{algorithm}
\begin{figure*}[t]
\begin{center}
\includegraphics[width=1.0\linewidth]{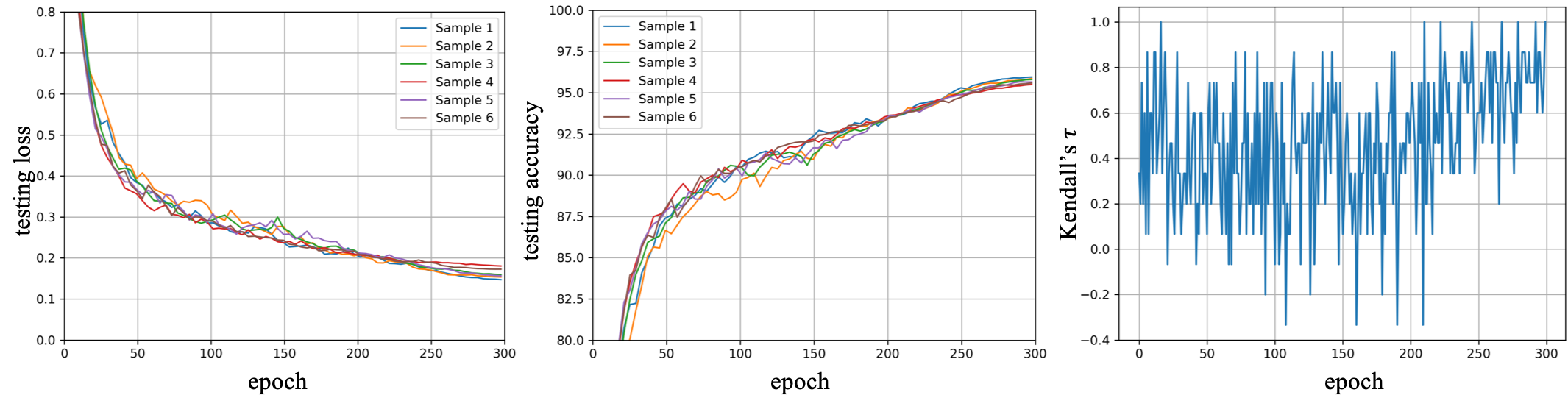}
\end{center}
   \caption{The test error (left), top 1 accuracy (middle), and Kendall's $\tau$ (right) of different architectures. The error and accuracy curves are entangled, since they are sampled from the same search space defined in Section~\ref{sec:3}. Therefore, we further calculate the Kendall's $\tau$ between every epoch and the final result. Note that the Kendall's $\tau > 0$ can be considered as a high value, which means more than half of the rankings are consistent.}
\label{fig:hypithesis_ev}
\end{figure*}
\section{Experiment}\label{experiment}
In this section, we first conduct some experiments on the CIFAR-10 to demonstrate the proposed hypothesis. Then, we compare our method with state-of-the-art methods on both search effectiveness and efficiency on two widely-used classification datasets including CIFAR-10 and ImageNet.
\subsection{Experiment Settings}
\subsubsection{Datasets}
We follow most NAS works \cite{liu2018darts,cai2018path,zoph2018learning,liu2018progressive} in their experiment datasets and evaluation metrics. In particular, we conduct most experiments on CIFAR-10 \cite{krizhevsky2010convolutional} which has $50,000$ training images and $10,000$ testing images. In architecture search, we randomly select $5,000$ images in the training set as the validation set to evaluate the architecture. The color image size is $32 \times 32$ with $10$ classes. All the color intensities of the images are normalized to $[-1, +1]$. To further evaluate the generalization, after discovering a good cell on CIFAR-10, the architecture is transferred into a deeper network, and therefore we also conduct classification on ILSVRC 2012 ImageNet \cite{russakovsky2015imagenet}. This dataset consists of $1,000$ classes, which has 1.28 million training images and $50,000$ validation images. Here we consider the \emph{mobile} setting where the input image size is $224 \times 224$ and the number of multiply-add operations in the model is restricted to be less than 600M.
\subsubsection{Implementation Details}\label{sec:details}
In the search process, according to the hypothesis, the layer number is irrelevant to the evaluation of a cell structures. We therefore consider in total $L=6$ cells in the network, where the reduction cells are inserted in the second and third layers, and $4$ nodes for a cell. The network is trained for 100 epoches, with a batch size as 512 (due to the shallow network and few operation sampling), and the initial number of channels as 16. We use SGD with momentum to optimize the network weights $w$, with an initial learning rate of 0.025 (annealed down to zero following a cosine schedule), a momentum of 0.9, and a weight decay of $3 \times 10^{-4}$. The learning rate of the multinomial parameters is set to 0.01. The search takes only 4 GPU hours with only one NVIDIA GTX 1080Ti on CIFAR-10. 

In the architecture evaluation step, the experimental setting is similar to \cite{liu2018darts,zoph2018learning,pham2018efficient}. A large network of 20 cells is trained for 600 epochs with a batch size of 96, with additional regularization such as cutout \cite{devries2017improved}, and path dropout of probability of 0.3 \cite{liu2018darts}. All the experiments and models of our implementation are in PyTorch \cite{paszke2017automatic}. 

On ImageNet, we keep the same search hyper-parameters as on CIFAR-10. In the training procedure, we follow previous NAS methods \cite{liu2018darts,zoph2018learning,pham2018efficient} with the same experimental settings. The network is trained for 250 epochs with a batch size of 512, a weight decay of $3 \times 10 ^{-5}$, and an initial SGD learning rate of 0.1 (decayed by a factor of 0.97 in every epoch).

\subsubsection{Baselines}
We compare our method with both human designed networks and other NAS networks. The manually designed networks include ResNet \cite{he2016deep}, DenseNet \cite{huang2017densely} and SENet \cite{hu2018squeeze}. For NAS networks, we classify them according to different search methods, such as RL (NASNet \cite{zoph2018learning}, ENAS \cite{pham2018efficient} and Path-level NAS \cite{cai2018path}), evolutional algorithms (AmoebaNet \cite{real2018regularized}), Sequential Model Based Optimization (SMBO) (PNAS \cite{liu2018progressive}), and gradient-based (DARTS \cite{liu2018darts}).  We further compare our method under the mobile setting on ImageNet to demonstrate the generalization. The best architecture generated by our algorithm on CIFAR-10 is transferred to ImageNet, which follows the same experimental setting as the works mentioned above. Since our algorithm takes less time and memory, we also directly search on ImageNet, and compare it with another similar baseline (low computation consumption) of proxy-less NAS \cite{cai2018proxylessnas}.
\begin{table*}[]
\begin{center}
\setlength{\tabcolsep}{7mm}{
\begin{tabular}{lcccc}
\toprule[1pt]
\multirow{2}{*}{\textbf{Architecture}} & \textbf{Test Error} & \textbf{Params} & \textbf{Search Cost} & \textbf{Search}        \\
                                       & \textbf{(\%)}       & \textbf{(M)}    & \textbf{(GPU days)} & \textbf{Method}        \\ \hline
ResNet-18 \cite{he2016deep}                              & 3.53                & 11.1            & -                    & manual                 \\
DenseNet \cite{huang2017densely}                      & 4.77                & \textbf{1.0}             & -                    & manual                 \\
SENet \cite{hu2018squeeze}                                  & 4.05                & 11.2            & -                    & manual                 \\ \hline
NASNet-A \cite{zoph2018learning}                       & 2.65                & 3.3             & 1800               & RL                     \\
AmoebaNet-A \cite{real2018regularized}                     & 3.34                & 3.2             & 3150               & evolution              \\
AmoebaNet-B \cite{real2018regularized}                     & 2.55                & 2.8             & 3150               & evolution              \\
PNAS \cite{liu2018progressive}                                   & 3.41                & 3.2             & 225                & SMBO \\
ENAS \cite{pham2018efficient}                                  & 2.89                & 4.6             & 0.5                   & RL                     \\
Path-level NAS \cite{cai2018path}                                  & 2.49                & 5.7             & 8.3                  & RL                     \\
DARTS(first order) \cite{liu2018darts}            & 2.94                & 3.1             & 1.5                   & gradient-based         \\
DARTS(second order) \cite{liu2018darts}            & 2.83                & 3.4             & 4                   & gradient-based         \\ 
Random Sample \cite{liu2018darts}            & 3.49                & 3.1             & -                   & -         \\
\hline
\textbf{MdeNAS (Ours)}                          & \textbf{2.55}       & 3.61   & \textbf{0.16}           & \textbf{MDL}           \\ \bottomrule[1pt]
\end{tabular}}
\end{center}
\caption{Test error rates of our discovered architecture, human-designed network and other NAS architectures on CIFAR-10. To be fair, we select the architectures and results with similar parameters ($<$ 10M) and training conditions (same epochs and regularization).}
\label{tab:cifar_results}
\end{table*}
\subsection{Evaluation of the Hypothesis}
We first conduct experiments to verify the correctness of the proposed performance ranking hypothesis. To get some intuitive sense of the hypothesis, we introduce the Kendall rank correlation coefficient, \emph{a.k.a.} Kendall's $\tau$ \cite{abdi2007kendall}. Given two different ranks of $m$ items, the Kendall's $\tau$ is computed as follows:
\begin{equation}\label{Kendall_tau}
\tau = \frac{P-Q}{P+Q},
\end{equation}
where $P$ is the number of pairs that are concordant (in the same order in both rankings) and $Q$ denotes the number of pairs that are discordant (in the reverse order). $\tau \in [-1, 1]$, with 1 meaning the rankings are identical and -1 meaning a rank is in reverse of another. The probability of a pair in two ranks being consistent is $p_{\tau} = \frac{\tau + 1}{2}$. Therefore, a $\tau = 0$ means that $50\%$ of the pairs are concordant.  

We randomly sample different network architectures  in the search space, and report the loss, accuracy and Kendall's $\tau$ of different epochs on the testing set. The performance ranking in every epoch is compared with the final performance ranking of different network architectures. As illustrated in Fig.~\ref{fig:hypithesis_ev}, the accuracy and loss are hardly distinguished due to the homogeneity of the sampled networks, \emph{i.e.,} all the networks are generated from the same space. On the other hand, the Kendall coefficient keeps a high value ($\tau>0, p_{\tau}>0.5$) in most epochs, generally approaching 1 as the number of epochs increases. It indicates that the architecture evaluation ranking has highly convincing probabilities in every epoch and generally becomes more close to the final ranking. Note that, the mean value of Kendall's $\tau$ for each epoch is 0.474. Therefore, the hypothesis holds with a probability of 0.74. Moreover, we discover that the combination of the hypothesis with the multinomial distribution learning can enhance each other. The hypothesis guarantees the high expectation when selecting a good architecture, and the distribution learning decreases the probability of sampling a bad architecture.
%

\begin{figure}
\begin{center}
\includegraphics[width=1.0\linewidth]{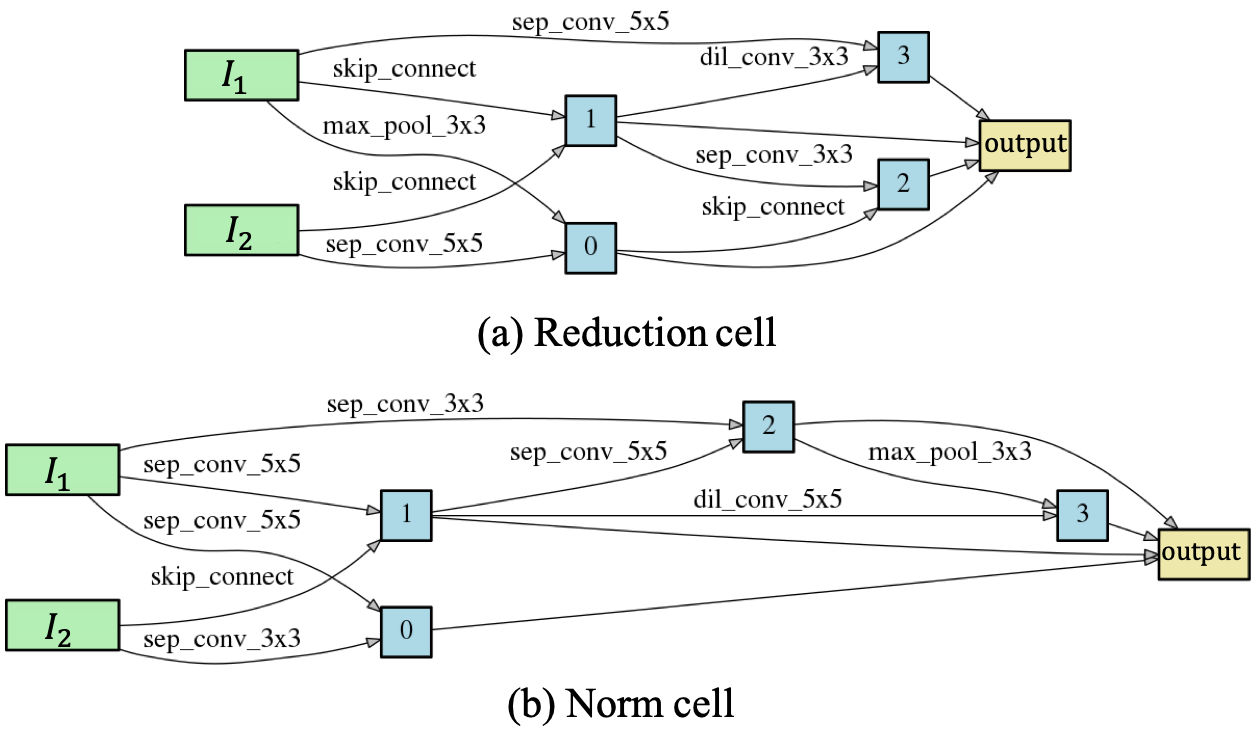}
\end{center}
   \caption{Detailed structure of the best cells discovered on CIFAR-10. The definition of the operations on the edges is in Section~\ref{sec:3.1}. In the reduction cell (up) the stride of operations on 2 input nodes is 2, and in the norm cell (down), the stride is 1. }
\label{fig:search_architecture}
\end{figure}

\begin{table*}[]
\begin{center}
\setlength{\tabcolsep}{5mm}{
\begin{tabular}{lccccc}
\toprule
\multirow{2}{*}{\textbf{Architecture}} & \multicolumn{2}{c}{\textbf{Accuracy (\%)}} & \textbf{Params} & \textbf{Search Cost} & \textbf{Search} \\ \cline{2-3}
                                       & \textbf{Top1}        & \textbf{Top5}       & \textbf{(M)}    & \textbf{(GPU days)} & \textbf{Method} \\ \hline
MobileNetV1 \cite{howard2017mobilenets}                            & 70.6                 & 89.5                & 6.6             & -                    & manual          \\
MobileNetV2 \cite{sandler2018mobilenetv2}                            & 72.0                 & 91.0                & 3.4             & -                    & manual          \\
ShuffleNetV1 2x (V1) \cite{zhang2018shufflenet}                  & 70.9                 & 90.8                & $\sim$5         & -                    & manual          \\
ShuffleNetV2 2x (V2) \cite{ma2018shufflenet}                   & 73.7                 & -                   & $\sim$5         & -                    & manual          \\ \hline
NASNet-A \cite{zoph2018learning}                               & 74.0                 & 91.6                & 5.3             & 1800                 & RL              \\
AmoebaNet-A \cite{real2018regularized}                           & 74.5                 & 92.0                & 5.1             & 3150                 & evolution       \\
AmoebaNet-C \cite{real2018regularized}                           & 75.7                 & 92.4                & 6.4             & 3150                 & evolution       \\
PNAS \cite{liu2018progressive}                                  & 74.2                 & 91.9                & 5.1             & 225                  & SMBO            \\
DARTS \cite{liu2018darts}                                 & 73.1                 & 91.0                & 4.9             & 4                    & gradient-based  \\ \hline
MdeNAS (Ours)                           &            74.5          &          92.1           & 6.1             & 0.16                  & MDL             \\         \bottomrule
\end{tabular}}
\end{center}
\caption{Comparison with state-of-the-art image classification methods on ImageNet with the mobile setting. All the NAS networks are searched on CIFAR-10, and then directly transferred to ImageNet.}
\label{tab:imagenet_results}
\end{table*}

\subsection{Results on CIFAR-10}
We start by finding the optimal cell architecture using the proposed method. In particular, we first search neural architectures on an over-parameterized network, and then we evaluate the best architecture with a deeper network. To eliminate the random factor, the algorithm is run for several times. We find that the architecture performance is only slightly different with different times, as well as comparing to the final performance in the deeper network ($<$0.2), which indicates the stability of the proposed method. The best architecture is illustrated in Fig.~\ref{fig:search_architecture}.

The summarized results for convolutional architectures on CIFAR-10 are presented in Tab.~\ref{tab:cifar_results}. It is worth noting that the proposed method outperforms the state-of-the-art \cite{zoph2018learning,liu2018darts}, while with extremely less computation consumption (only 0.16 GPU days $<<$ 1,800 in \cite{zoph2018learning}). Since the performance highly depends on different regularization methods (\emph{e.g.}, cutout \cite{devries2017improved}) and layers, the network architectures are selected to compare equally under the same settings. Moreover, other works search the networks using  either differential-based or black-box optimization. We attribute our superior results based on our novel way to solve the problem with distribution learning, was well as the fast learning procedure: The network architecture can be directly obtained from the distribution when the distribution converges. On the contrary, previous methods \cite{zoph2018learning} evaluate architectures only when the training process is done, which is highly inefficient. Another notable phenomena observed in Tab.~\ref{tab:cifar_results} is that, even with randomly sampling in the search space, the test error rate in \cite{liu2018darts} is only 3.49\%, which is comparable with the previous methods in the same search space. We can therefore reasonable conclude that, the high performance in the previous methods is partially due to the good search space. At the same time, the proposed method quickly explores the search space and generates a better architecture. We also report the results of hand-crafted networks  in Tab.~\ref{tab:cifar_results}. Clearly, our method shows a notable enhancement, which indicates its superiority in both resource consumption and test accuracy.

\begin{table}[]
\begin{center}
\setlength{\tabcolsep}{1.2mm}{
\begin{tabular}{lccc}
\toprule
\multirow{2}{*}{\textbf{Model}} & \multirow{2}{*}{\textbf{Top-1}} & \textbf{Search time} & \multirow{2}{*}{\textbf{GPU latency}} \\
                                &                                 & \textbf{GPU days}    &                                       \\ \hline
MobileNetV2                     & 72.0                            & -                    & 6.1ms                                 \\
ShuffleNetV2                    & 72.6                            & -                    & 7.3ms                                 \\ \hline
Proxyless (GPU) \cite{cai2018proxylessnas}                 & 74.8                            & 4                    & 5.1ms                                 \\
Proxyless (CPU)  \cite{cai2018proxylessnas}                 & 74.1                            & 4                    & 7.4ms                                 \\ \hline
MdeNAS (GPU)                     &                 75.2                & 2                    & 4.9ms                                 \\
MdeNAS (CPU)                     &                  74.1               & 2                    &          7.1ms                            
\\ \bottomrule
\end{tabular}}
\end{center}
\caption{Comparison with state-of-the-art image classification on ImageNet with the mobile setting. The networks are directly searched on ImageNet with the MobileNetV2 \cite{sandler2018mobilenetv2} backbone.}
\label{tab:search_imagenet}
\end{table}

\subsection{Results on ImageNet}
We also run our algorithm on the ImageNet dataset \cite{russakovsky2015imagenet}. Following existing works, we conduct two experiments with different search datasets, and test on the same dataset. As reported in Tab.~\ref{tab:cifar_results}, the previous works are time consuming on CIFAR-10, which is impractical to search on ImageNet. Therefore, we first consider a transferable experiment on ImageNet, \emph{i.e.}, the best architecture found on CIFAR-10 is directly transferred to ImageNet, using two initial convolution layers of stride 2 before stacking 14 cells with scale reduction (reduction cells) at 1, 2, 6 and 10. The total number of flops is decided by choosing the initial number of channels. We follow the existing NAS works to compare the performance under the \emph{mobile} setting, where the input image size is $224 \times 224$ and the model is constrained to less than 600M FLOPS. We set the other hyper-parameters by following \cite{liu2018darts,zoph2018learning}, as mentioned in Sec.~\ref{sec:details}. The results in Tab.~\ref{tab:imagenet_results} show that the best cell architecture on CIFAR-10 is transferable to ImageNet. Note that, the proposed method achieves comparable accuracy with state-of-the-art methods, while using much less computation resource.

The extremely minimal time and GPU memory consumption makes our algorithm on ImageNet feasible. Therefore, we further conduct a search experiment on ImageNet. We follow \cite{cai2018proxylessnas} to design network setting and the search space. In particular, we allow a set of mobile convolution layers with various kernels $\{3,5,7\}$ and expanding ratios $\{1,3,6\}$. To further accelerate the search, we directly use the network with the CPU and GPU structure obtained in \cite{cai2018proxylessnas}. In this way, the zero and identity layer in the search space is abandoned, and we only search the hyper-parameters related to the convolutional layers. The results are reported in Tab.~\ref{tab:search_imagenet}, where we have found that our MdeNAS achieves superior performance compared to both human-designed and automatic architecture search methods, with less computation consumption. 

\section{Conclusion}
In this paper, we have presented MdeNAS, the first distribution learning-based architecture search algorithm for convolutional networks.   Our algorithm is deployed based on a novel \emph{performance rank hypothesis} that is able to further reduce the search time which compares the architecture performance in the early training process. Benefiting from our hypothesis, MdeNAS can drastically reduce the computation consumption while achieving excellent model accuracies on CIFAR-10 and ImageNet. Furthermore, MdeNAS can directly search on ImageNet,  which outperforms the human-designed networks and other NAS methods.

\paragraph{Acknowledgements.}
This work is supported by the National Key R\&D Program (No.2017YFC0113000, and No.2016YFB1001503), Nature Science Foundation of China (No.U1705262, No.61772443, and No.61572410), Post Doctoral Innovative Talent Support Program under Grant BX201600094, China Post-Doctoral Science Foundation under Grant 2017M612134, Scientific Research Project of National Language Committee of China (Grant No. YB135-49), and Nature Science Foundation of Fujian Province, China (No. 2017J01125 and No. 2018J01106).
{
\small
\bibliographystyle{ieee_fullname}
\bibliography{ICCV2019_1865}
}

\end{document}